\documentclass[10pt,A4paper]{article}

\usepackage{arxiv}
\usepackage{authblk}
\usepackage{latexsym}
\usepackage{graphicx}
\usepackage{multicol,multirow}
\usepackage{amsmath,amssymb,amsfonts}
\usepackage{mathrsfs}
\usepackage{amsthm}
\usepackage{apacite}
\usepackage{rotating}
\usepackage{appendix}
\usepackage[authoryear]{natbib}
\usepackage{ifpdf}
\usepackage[T1]{fontenc}
\usepackage{times}
\usepackage{sourcesanspro}
\usepackage{newtxmath}
\usepackage{textcomp}%
\usepackage{xcolor}%
\usepackage{hyperref}
\graphicspath{ {./Figures_new/} }
\usepackage{subfigure}
\usepackage{comment}
\usepackage{float}
\usepackage{url}
\usepackage{color}

\newcommand\norm[1]{\left\lVert#1\right\rVert}
\DeclareMathOperator*{\argmin}{arg\,min}

\title{Learning stable reduced-order models for hybrid twins}

\author[1,2,4]{Abel Sancarlos}
\author[2]{Morgan Cameron}
\author[3]{Jean-Marc Le Peuvedic}
\author[2]{Juliette Groulier}
\author[2]{Jean-Louis Duval}
\author[4]{Elias Cueto}
\author[1,2]{Francisco Chinesta}

\affil[1]{{\small PIMM lab and ESI Group Chair. ENSAM Institute of Technology, Paris, France.}}
\affil[2]{{\small ESI Group. 3bis, rue Saarinen, 94528 Rungis CEDEX, France.}}
\affil[3]{{\small Dassault Aviation. 78 Quai Marcel Dassault. 92210 Saint-Cloud, France}}
\affil[4]{{\small Aragon Institute of Engineering Research. Universidad de Zaragoza. Zaragoza, Spain.}}

\begin{document}

\maketitle

\begin{abstract}
The concept of ``Hybrid Twin''  (HT) has \textcolor{black}{recently} received a growing interest thanks to the availability of powerful machine learning techniques. This twin concept combines physics-based models within a model-order reduction framework---to obtain real-time feedback rates---and data science. Thus, the main idea of the HT is to develop on-the-fly data-driven models to correct possible deviations between measurements and physics-based model predictions. This paper is focused on the computation of stable, fast and accurate corrections in the Hybrid Twin framework. Furthermore, regarding the delicate and important problem of stability, a new approach is proposed, introducing several sub-variants and guaranteeing a low computational cost as well as the achievement of a stable time-integration.\end{abstract}

\section{Introduction}
%
%

\begin{figure}
\centering
\includegraphics[width=0.9\linewidth]{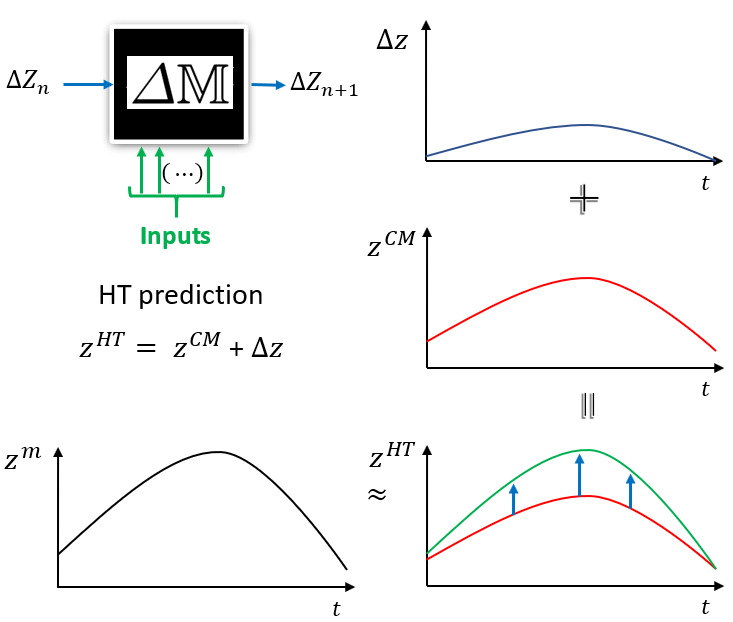}
\caption{ Diagram illustrating the HT concept. The HT is able to correct the discrepancy between the coarse model (CM) and the pseudo-experimental data (PED, denoted by a superscript $m$). Its prediction $z$ is here denoted by a superscript $HT$ whereas the enrichment model is denoted by $\Delta z$. The superscript $CM$ refers to the ``coarse model'', and $\Delta\mathbb M$ is the model correction. Both concepts are introduced in Section \ref{Sec:SystemModeling}\label{HTconcept}}
\end{figure}

The Hybrid Twin (HT) paradigm is a powerful tool to make better predictions, increase control performance or improve decision-making \citep{Chi,HTclara}. The main idea, see Fig.~\ref{HTconcept}, is to develop on-the-fly data-driven models to correct the gap between data (i.e., measurements) and model predictions. In other words, there are two main ingredients of a HT: 
\begin{itemize}
\item The first one is to enrich physics description with data. 
\item The second one is to accelerate physics-based models using Model Order Reduction (MOR) techniques, as in  \citet{MOR_Paco}, \citet{MOR_PGD}, or \citet{Abel_Motors1}.
\end{itemize}
In any case, when addressing dynamical systems in the HT framework, it is important to guarantee the stability of the system when adding corrections to the physical model. It is worth noting that this is an important issue, because sometimes the best model, computed with state-of-the-art algorithms, completely fails to obtain a stable \textcolor{black}{time-integrator}. For example, when considering a linear dynamical model by the Dynamic Mode Decomposition (DMD) approach \citep{Sch,DMD_book}, the feasible region constrained by the stability condition is nonconvex \citep{key_paper_stability}, and no general methodology exists to solve it.

For this reason, this work proposes a new, fast and efficient methodology, covering several sub-variants and guaranteeing a low computational cost as well as the achievement of a stable dynamical system. This technique will therefore be used to add a stable correction term \textcolor{black}{into} the HT concept.

The methodology is tested for an industrial case detailed in Section \ref{Sec:typedata}, where an excellent agreement was observed when employing the HT approach.

The work is organized as follows: in Section \ref{Sec:SystemModeling}, the system modeling with the HT concept is presented and compared to the direct (so to speak, from scratch) data-driven approach. An alternative approach that benefits from transfer learning, for instance, can be found in \citet{Vinuesa2}. Then, in Section \ref{Sec:Stable}, the proposed technique to obtain stable systems is described as well as the sub-variants to deal with high-dimensional systems. Finally, in Sections \ref{Sec:Results} and \ref{Sec:Conclusions}, the results and general conclusions of the present work are discussed, respectively.

\section{System modeling} \label{Sec:SystemModeling}
  
In what follows, we consider the system as described by a vector $\mathbf z \in \mathbb R^\mathtt D$ (with $\mathtt D$ the number of variables involved in the system evolution). The state (snapshot of the system) at time $t_n = n\Delta t$ is stored at vectors $\mathbf Z_n$, with $n \geq 0$, with $\mathbf Z_0$ assumed known. {In addition, $\mathtt d$ control parameters are considered, giving rise to a parametric space $\boldsymbol \mu \in \mathbb R^\mathtt d$.}

We assume the existence of a model $\mathbb M^c(\mathbf z,\boldsymbol \mu)$ which we refer to as coarse, since we assume that some form of enrichment is necessary. Often, this model is physics-based, it arises from the corresponding PDEs governing the problem and can contain non-linearities. Other times, it is based on MBSE (Model Based System Engineering) modelling. This approximate representation of the reality is sought to be computable under real-time constraints. These constraints depend on the context and can range from some seconds to the order of milliseconds. If the complexity of the model does not allow to obtain such a response under constraints, model order reduction techniques constitute an appealing alternative. These, that can be linear or not, allow us to timely integrate the state of the system.

Because of the simplifying hypotheses involved in the construction of the model $\mathbb M^c$, it is expected that model predictions $\mathbf Z_n^c$ differ from measurements to some extent, i.e., {$\| \mathbf Z_n - \mathbf Z_n^c\| > \epsilon$}, for most time steps $n$, thus needing for a correction.


\subsection{Extracting the model of the system {from scratch}}\label{M1}

\textcolor{black}{Several routes exist to construct a model for a given dynamical system. The first one} consists in  performing a completely data-driven approach from experimental measurements $(\mathbf Z_n, \boldsymbol \mu_n)$, $n =  0, \ 1, \ldots$. A valuable option is to consider the so-called dynamic mode decomposition (DMD)  to extract a matrix model of a discrete linear system \citep{Sch}. As suggested by several works, many different systems can be well approximated using this approach \citep{DMDfluids1,DMDfluids2}. \textcolor{black}{If problems arise due to complex system behaviours, the procedure of extending the state vector to a higher dimensional space can often solve the problem} \citep{DMD_book,Vinuesa1}.

A second alternative could be a variant of the technique presented in \citep{sancarlos2020rom}, that consists in grouping all the states close to $\mathbf Z_n$ and the control parameters close to $\boldsymbol \mu_n$ into a set $\mathcal S_n$. For the sake of clarity,  in what follows $\mathcal S$ will refer to one of these generic sets. A linear model for the set $\mathcal S$, denoted $\mathbf{M_{\mathcal S}}$, {which in this case is simply a matrix,} is extracted from 
\begin{equation} \label{eq:ols}
\mathbf{M_{\mathcal S}} = \argmin_{\mathbf{N_{\mathcal S}}} \norm{ \mathbf X_{1} - \mathbf{N_{\mathcal S}} \tilde{\mathbf{X}}_0 }_F^2, 
\end{equation}
with
\begin{equation*}
\tilde{\mathbf{Z}}_n
=
\begin{bmatrix} 
\mathbf{Z}_{n} \\
\boldsymbol \mu_{n}
\end{bmatrix},
\end{equation*}
\begin{equation*}
\tilde{\mathbf{X}}_0=
\begin{bmatrix} 
\tilde{\mathbf{Z}}_0, & \tilde{\mathbf{Z}}_1 ,& \ldots ,& \tilde{\mathbf{Z}}_{n_{s-1}} 
\end{bmatrix},
\end{equation*}
\begin{equation*}
\mathbf{X}_1=
\begin{bmatrix} 
\mathbf{Z}_1, & \mathbf{Z}_2, & \ldots ,& \mathbf{Z}_{n_s} 
\end{bmatrix},
\end{equation*}
if stability problems do not arise. Note that in this case the model is composed of local matrices defining different linear maps in each set.
A reduced version of this approach was extensively discussed in \citep{Rei}. 

Quite often, \textcolor{black}{an issue} can appear because of the difficulties to learn stable models when constructing these dynamical systems. For this reason, in Section \ref{Sec:Stable}, we propose a new methodology when using these techniques to guarantee that the obtained systems remain stable.

\subsection{Enriching a physics-based model within the Hybrid Twin framework}\label{M2}

Constructing a model $\mathbb M$ of the physical system from scratch is not the most valuable route as discussed in our former works, cf. \citep{Chi}. Purely data-driven, black-box models are not popular in industry, due to the lack of interpretability and guaranteed error estimators. Thus, a more valuable option consists of constructing corrections to physics-based models---if these provide unsatisfactory results---from an additive correction of this coarse prediction.

In fact, since the coarse model is expected to perform reasonably well for predicting the state of the system, bias will in general remain reasonably small. If this is true, the correction model will be much less nonlinear, and it will accept a more accurate description from the same amount of data.

Thus, we define the correction contribution $\mathbf C_{n}$ (or, equivalently, the model enrichment) as:
\begin{equation}
\mathbf Z_n 
- 
\mathbf Z^c_n 
= 
\mathbf C_n,
\end{equation}
where $\mathbf Z_n^c$ refers the model prediction.

Taking as a proof of concept a dynamic linear system with \textcolor{black}{control inputs}, the correction term is \textcolor{black}{searched as}:
\begin{equation}
\mathbf C_{n+1} =  \mathbf W \mathbf C_n + \mathbf V \boldsymbol \mu_n,
\end{equation}
where $\mathbf W$ and $\mathbf V$ are the matrices defining the time evolution of the correction term. 
Thus, the total response of the system is finally predicted by using
\begin{equation}
\mathbf Z_{n+1} 
\approx
\mathbf Z^c_{n+1} + \mathbf W \mathbf C_n + \mathbf V \boldsymbol \mu_n.
\end{equation} 

By taking into consideration that $\mathbf Z^c$ is stable and integrated independently of the correction term, the stability condition of the system \textcolor{black}{should apply just on} the correction term.

We propose an approach to guarantee the construction of stable systems with low computational cost in the next section. \textcolor{black}{In addition, this technique can be used either to build models from scratch or the correction term of the HT approach.}

\section{Efficiently learning stable linear dynamical systems and DMDc models.} \label{Sec:Stable}

Frequently, difficulties to learn stable models arise when learning linear dynamical systems, specially when dealing with high-dimensional data. 
This is an important issue to deal with because of the growing importance of the data-driven approximations. For instance, it is usual to search the best linear approach of a set of high-dimensional data to make, for example, fast predictions of the system or to develop control strategies. In fact, there is a growing success of techniques such as the DMD  to discover dynamical systems from high-dimensional data \citep{Sch,DMD_book}. This success steams from its capability of providing an accurate decomposition of a complex system into spatiotemporal coherent structures while constructing the model dynamics evolving on a low-rank subspace.

However, when operating in the above scenarios, sometimes the best model computed with state-of-the-art algorithms fails to obtain a stable \textcolor{black}{time-integrat\-ion}. In fact, for a given a set of data, the problem to guarantee a stable system is defined by Eq. \eqref{eq:rhocond} below but, unfortunately, the feasible region constrained by the spectral radius is nonconvex  and no general methodology exists to solve it \citep{key_paper_stability}. Moreover, if a fast procedure is needed to obtain a correction model in a real-time application, this problem is further exacerbated.

For these reasons, this paper proposes a methodology to compute a stable model for a given dataset at low computational cost. The strategy is discussed for  a discrete linear system, for a DMD model and for a DMD with control (DMDc) model. In addition, other strategies are discussed.

Let us assume the following dynamical systems defined in Eqs \eqref{eq:DS_noI} and \eqref{eq:DS_wI}, the first one without considering inputs,
\begin{equation} \label{eq:DS_noI}
\mathbf{Z}_{n+1} =
\mathbf{M}
\mathbf{Z}_{n},
\end{equation}
and the second one considering inputs,
\begin{equation} \label{eq:DS_wI}
\mathbf{Z}_{n+1} =
\begin{bmatrix} 
\mathbf{M} & \mathbf{N} 
\end{bmatrix}
\begin{bmatrix} 
\mathbf{Z}_{n} \\
\boldsymbol \mu_{n}
\end{bmatrix}
=
\tilde{\mathbf{M}}
\tilde{\mathbf{Z}}_{n},
\end{equation}
where $\mathbf{M}$ and $\mathbf{N}$ are the matrices defining the time evolution of the system.
To guarantee stability in the above systems, the following condition must be satisfied:
\begin{equation} \label{eq:rhocond}
\rho(\mathbf{M}) \leq 1,
\end{equation}
where $\rho(\cdot)$ denotes the spectral radius.

Therefore, in relation to Eq. \eqref{eq:DS_wI}, the solution minimizes
\begin{equation} \label{minDMok}
\begin{aligned}
\norm{\mathbf{X}_1-
\tilde{\mathbf{M}} \tilde{\mathbf{X}}_0
}_F^2 &    &
\text{s. t.}
&    & \rho(\mathbf{M}) \leq 1,
\end{aligned}
\end{equation}
with
\begin{equation*}
\tilde{\mathbf{X}}_0=
\begin{bmatrix} 
\tilde{\mathbf{Z}}_0, & \tilde{\mathbf{Z}}_1 ,& \ldots ,& \tilde{\mathbf{Z}}_{n_{s-1}} 
\end{bmatrix},
\end{equation*}
\begin{equation*}
\mathbf{X}_1=
\begin{bmatrix} 
\mathbf{Z}_1, & \mathbf{Z}_2, & \ldots ,& \mathbf{Z}_{n_s} 
\end{bmatrix},
\end{equation*}
where $n_s$ is the number of different snapshots for the training and the matrices $\tilde{\mathbf{X}}_0$ and $\mathbf{X}_1$ contain the data to construct the model. In them, each column corresponds to a snapshot of the system at a given time instant.

Unfortunately, as already said, the feasible region constrained by $\rho(\mathbf{\mathbf{M}}) \leq 1$ is nonconvex  and no general methodology exists to solve it \citep{key_paper_stability}. This can lead to the problems already discussed where an unstable model is obtained or, in other cases, simply an extremely bad model is extracted due to a failure of the optimization methodology employed.

The proposed approach, which can always guarantee  the creation of a stable model, is based on an observation of the following inequality, which is satisfied by any matrix norm:
\begin{equation} \label{eq:proof1}
\rho(\mathbf{A}) \leq \norm{\mathbf{A}}.
\end{equation}

\textcolor{black}{To prove it, let} $\omega$ be an eigenvalue of $\mathbf A$, and let $\mathbf x \neq 0$ be a corresponding eigenvector. From $\mathbf A \mathbf x = \omega \mathbf x$, we have:
\begin{equation*}
\mathbf A \mathbf X = \omega \mathbf X,
\end{equation*}
where $\mathbf X = [\mathbf x, \cdots, \mathbf x]$.

It follows that
\begin{equation*}
\norm{\omega \mathbf X}
=
\lvert \omega \rvert \norm{\mathbf X},
\end{equation*}
and taking into account that
\begin{equation*}
\norm{\mathbf A \mathbf X}
\leq
\norm{\mathbf A} \norm{\mathbf X},
\end{equation*}
it follows that 
\begin{equation*}
\lvert \omega \rvert \norm{\mathbf X}
\leq
\norm{\mathbf A} \norm{\mathbf X}.
\end{equation*}

Simplifying the above expression by $\norm{\mathbf X}$ ($> 0$) gives:
\begin{equation*}
\lvert \omega \rvert 
\leq
\norm{\mathbf A},
\end{equation*}
that taking the maximum over all eigenvalues $\omega$ gives the desired proof. 

Taking for the reasoning the following induced norm,
\begin{equation*} \label{eq:proof2}
\norm{\mathbf{A}}_1
=
\max_{1 \leq j \leq m_1} 
\sum_{i=1}^{m_2} \lvert a_{ij} \rvert,
\end{equation*}
where 
$m_1$ is the number of columns, and $m_2$ is the number of rows---which is simply the maximum absolute column sum of the matrix. It can be observed that by decreasing the absolute value of the matrix coefficients, a smaller matrix norm is obtained, and if it is decreased sufficiently, a smaller $\rho(\mathbf{A})$ is got, because of Eq.~\eqref{eq:proof1}.

Therefore, the idea to obtain a stable system is to shrink the matrix coefficients of $\mathbf{M}$. In fact, we propose to do so by using the ridge regression \citep{Key_Stats}, also known as a special case of the Tikhonov regularization. Many advantages can be obtained from this choice. For instance, a closed mathematical expression is obtained. This implies that there is no need to use complex optimization procedures (that can fail to converge). In addition, there is a low added computational cost when changing the Ordinary Least Squares (OLS) problem to the ridge one.

This way, we proved a new feature and use for the ridge regression. Ridge regression was employed in \textcolor{black}{certain} way for regression of dynamical systems, see e.g. \citet{randomDMD}, but just to use the classical function of ridge: to deal with ill-posed problems. Now, we have extended the employment of the technique to a broader problem area: the construction of stable dynamical systems.

To reformulate the resolution of the systems \eqref{eq:DS_noI} and \eqref{eq:DS_wI}, and then extend the procedure for the DMDc (a more general version of the DMD considering control inputs), two options are envisaged. The first one is solving the following problem:
\begin{multline}
\label{eq:ridge1}
 \hat{\mathbf{M}} = \argmin_{\mathbf{M}}
\Big\{
\norm{\mathbf{X}_1-\mathbf{M}  \mathbf{X}_0}_F^2
+
\lambda^2
\norm{\mathbf{M}}_F^2
\Big\}
\\
=
\argmin_{\mathbf{M}}
\norm{
\begin{bmatrix} 
\mathbf{X}_{1} & \mathbf{0}
\end{bmatrix}
-\mathbf{M}  
\begin{bmatrix} 
\mathbf{X}_{0} & \lambda \mathbf I
\end{bmatrix}
}_F^2
=
\argmin_{\mathbf{M}}
\norm{
\mathbf{\bar{X}}_{1}
-\mathbf{M}   
\mathbf{\bar{X}}_{0}
}_F^2,
\end{multline}
where:
\begin{equation*}
\mathbf{X}_0=
\begin{bmatrix} 
\mathbf{Z}_0, & \mathbf{Z}_1 ,& ... ,& \mathbf{Z}_{n_s-1} 
\end{bmatrix},
\end{equation*}
$\mathbf{\bar{X}}_{1}$ and $\mathbf{\bar{X}}_{0}$ are the augmented matrices, $\mathbf I$ is an identity matrix of size $\mathtt D \times \mathtt D$ and $\mathbf{0}$ is a zero matrix of size $\mathtt D \times \mathtt D$.

The solution of the above problem can be computed using the Moore-Penrose pseudoinverse, therefore:
\begin{equation*}\label{eq:ridge2}
\hat{\mathbf{M}} = \mathbf{\bar{X}}_{1} (\mathbf{\bar{X}}_{0})^{\dag},
\end{equation*}
where $\dag$ is the the Moore-Penrose pseudoinverse.

A second procedure to solve Eq.~\eqref{eq:ridge1} is to employ a ridge regression for each variable to be predicted. Concerning the system \eqref{eq:DS_wI},
\begin{multline}
\label{eq:ridge3}
 \hat{\mathbf{M}} = \argmin_{\tilde{\mathbf{M}}}
\Big\{
\norm{\mathbf{X}_1-\tilde{\mathbf{M}}  \tilde{\mathbf{X}}_0}_F^2
+
\lambda^2
\norm{\mathbf{\textcolor{black}{M}}}_F^2
\Big\}
\\
=
\argmin_{\tilde{\mathbf{M}}}
\norm{
\begin{bmatrix} 
\mathbf{X}_{1} & \mathbf{0}_1
\end{bmatrix}
-\tilde{\mathbf{M}}  
\begin{bmatrix} 
\mathbf{X}_0 & \lambda \mathbf I \\
\mathbf{U}_0 & \mathbf{0}_2
\end{bmatrix}
}_F^2
=
\argmin_{\tilde{\mathbf{M}}}
\norm{
\mathbf{\bar{X}}_{1}
-\tilde{\mathbf{M}}   
\mathbf{\bar{Y}}_0
}_F^2,
\end{multline}
where:
\begin{equation*}
\mathbf{U}_0=
\begin{bmatrix} 
\boldsymbol \mu_1, & \boldsymbol \mu_2 ,& \ldots ,& \boldsymbol \mu_{n_{s-1}} 
\end{bmatrix},
\end{equation*}
$\mathbf{\bar{X}}_{1}$ and $\mathbf{\bar{Y}}_0$ are the augmented matrices, $\mathbf I$ is an identity matrix of size $\mathtt D \times \mathtt D$, $\mathbf{0}_1$ is a zero matrix of size $\mathtt D \times \mathtt D$ and $\mathbf{0}_2$ is a zero matrix of size $\mathtt d \times \mathtt D$.

In addition, the same procedures used to solve system \eqref{eq:ridge1} can be used to solve \eqref{eq:ridge3}, either the Moore-Penrose pseudoinverse or the individual ridge regressions.

The above model can be extended in the context of high-dimensional systems when using the DMDc taking the formulation expressed in Eq.~\eqref{eq:ridge3}. To do that, we take the Singular Value Decomposition (SVD) of matrix $\mathbf{\bar{Y}}_0 = \tilde{\mathbf{\Xi}} \tilde{\mathbf{\Sigma}} \tilde{\mathbf{V}}^*$ (where the star symbol * indicates the conjugate transpose). Therefore:
\begin{equation*}
\tilde{\mathbf{M}} =
\mathbf{\bar{X}}_{1} (\mathbf{\bar{Y}}_0)^{\dag}
=
\mathbf{\bar{X}}_{1}  \tilde{\mathbf{V}} \tilde{\mathbf{\Sigma}}^{-1} \tilde{\mathbf{\Xi}}^*.
\end{equation*}

Approximations of the operators $\mathbf{M}$ and $\mathbf{N}$ can be found as follows:
\begin{equation*}
\tilde{\mathbf{M}} =
\begin{bmatrix} 
\mathbf{M}, & \mathbf{N} 
\end{bmatrix}
=
\begin{bmatrix} 
\mathbf{\bar{X}}_{1}  \tilde{\mathbf{V}} \tilde{\mathbf{\Sigma}}^{-1} \tilde{\mathbf{\Xi}}_1^*, 
& 
\mathbf{\bar{X}}_{1}  \tilde{\mathbf{V}} \tilde{\mathbf{\Sigma}}^{-1} \tilde{\mathbf{\Xi}}_2^* 
\end{bmatrix},
\end{equation*}
where $\tilde{r}$ is the truncation value of the SVD applied to decompose matrix $\mathbf{\bar{Y}}_0 = \tilde{\mathbf{\Xi}} \tilde{\mathbf{\Sigma}} \tilde{\mathbf{V}}^*$, 
$\tilde{\mathbf{\Xi}}^* = \begin{bmatrix} 
\tilde{\mathbf{\Xi}}_1^*, & \tilde{\mathbf{\Xi}}_2^* 
\end{bmatrix}$, and the sizes of $\tilde{\mathbf{\Xi}}_1^* $ and $\tilde{\mathbf{\Xi}}_2^* $ are $\tilde{r} \times \mathtt D$ and $\tilde{r} \times \mathtt d$ respectively.

For high-dimensional systems ($\mathtt D \gg 1$) a reduced-order approximation can be solved for instead, leading to a more tractable computational model. Thus, we look for a transformation to a lower-dimensional subspace on which the dynamics evolve.

The output space $\mathbf{\bar{X}}_{1}$ is chosen to find the reduced-order subspace. Consequently, the SVD of $\mathbf{\bar{X}}_{1}$ is defined as
\begin{equation*}
\mathbf{\bar{X}}_{1}
=
\mathbf{\Xi} \mathbf{\Sigma} \mathbf{V}^*,
\end{equation*}
where the truncation value for this SVD will be denoted as $r$. Please note that usually both SVDs will have different truncation values.

Then, by employing the change of coordinates $\mathbf{Z} = \mathbf{\Xi} \, \hat{\mathbf{Z}}$ (or equivalently $\hat{\mathbf{Z}} = \mathbf{\Xi}^* \, \mathbf{Z}$), the following reduced-order approximations can be obtained,
\begin{equation*}
\mathbf{Z}_{n+1}
=
\mathbf{M} \mathbf{Z}_{n} +
\mathbf{N} \boldsymbol \mu_{n},
\end{equation*}
\begin{equation*}
\mathbf{\Xi} \, \hat{\mathbf{Z}}_{n+1}
=
\mathbf{M} \, \mathbf{\Xi} \, \hat{\mathbf{Z}}_{n} +
\mathbf{N} \, \boldsymbol \mu_{n},
\end{equation*}
so that
\begin{equation} \label{eq:ridgerom}
\hat{\mathbf{Z}}_{n+1}
=
\hat{\mathbf{M}} \, \hat{\mathbf{Z}}_{n} +
\hat{\mathbf{N}} \, \boldsymbol \mu_{n},
\end{equation}
where:
\begin{equation*}
\hat{\mathbf{M}} 
=
\mathbf{\Xi}^*
\mathbf{\bar{X}}_{1}  \tilde{\mathbf{V}} \tilde{\mathbf{\Sigma}}^{-1} \tilde{\mathbf{\Xi}}_1^* \mathbf{\Xi},
\end{equation*}
\begin{equation*}
\hat{\mathbf{N}} 
=
\mathbf{\Xi}^*
\mathbf{\bar{X}}_{1}  \tilde{\mathbf{V}} \tilde{\mathbf{\Sigma}}^{-1} \tilde{\mathbf{\Xi}}_2^*,
\end{equation*}
and the sizes of $\hat{\mathbf{M}}$ and $\hat{\mathbf{N}}$ are $r \times r$ and $r \times \mathtt d$ respectively.


To select the penalty factor, when the standard procedure leads to a matrix that violates the stability condition, several options can be envisaged. Here, we propose to use the bisection method (which guarantees convergence toward the solution) or the regula falsi method (to speed up the process). Of course, faster algorithms can be used such as the Illinois algorithm, but our experience suggests that the former ones are enough in practice.

For example, if the bisection method is selected, the zero of the following function is sought:
\begin{equation} \label{eq:functionlambda}
f(\lambda) = \rho_{\text{desired}} - \rho(\mathbf{M(\lambda)}),
\end{equation}
where $\rho_{\text{desired}}$ is a chosen value very close to one, representing the target when the initial constructed model violates the stability condition.

Taking into consideration that the bracketing interval at step $k$ of the algorithm is $[\lambda_k^a, \lambda_k^b]$, then the Eq.~\eqref{eq:bisection} is employed to compute the new solution estimate for the penalty $\lambda_k^c$ at step $k$:
\begin{equation} \label{eq:bisection}
\lambda_k^c =
\frac{\lambda_k^a + \lambda_k^b}{2}.
\end{equation}

If $f(\lambda_k^c)$ is satisfactory, the iteration stops.
If this is not the case, the sign of $f(\lambda_k^c)$ is examinated and the bracking interval is updated for the following iteration so that there is a zero crossing within the new interval.

\section{Application to a dynamical system} \label{Sec:Results}

\subsection{System to model and types of data.}
\label{Sec:typedata}

The modeled system {corresponds to an air distribution system of an aircraft and} is characterized by eight variables defining the state of the system: six temperatures $T^i_n$, $i=1, 2, \ldots, 6$, and two pressures $p^j_n$, $j=1,2$. The model should also take into account three control variables $\mu^k_n$, $k=1, 2, 3$, for each time instant $n$.

{With the knowledge and experience of Dassault Aviation, two models} are constructed with the help of the  software Dymola \citep{Dymola}:
\begin{itemize}
\item A coarse model (CM). This model deliberately fails to provide accurate predictions due to over-simplification. It is important to note that in industry, this type of model is often physics-based (although this is not mandatory) but still requires an important computing time.
\item A high-fidelity model that will therefore be considered as the ground truth  { (GT), which is consequently still more time consuming. This model is going to emulate in this work the real state of the system.}
\end{itemize}
Due to confidentiality issues, Dassault Aviation simulated different flights with both models and provided three different types of pseudo-experimental data which are employed in the present work:
\begin{itemize}
\item The CM data. These data correspond to the predictions of the CM for the given set of simulated flights. 
\item The GT data. These data correspond to the predictions of the GT model. It will be considered in the present work for evaluation purposes.
\item Pseudo-experimental data (PED). \textcolor{black}{A white} noise is added artificially to the GT data. Consequently, these data will emulate experimental measurements including experimental errors.
\end{itemize}

\begin{figure}
\centering
\includegraphics[width=\linewidth]{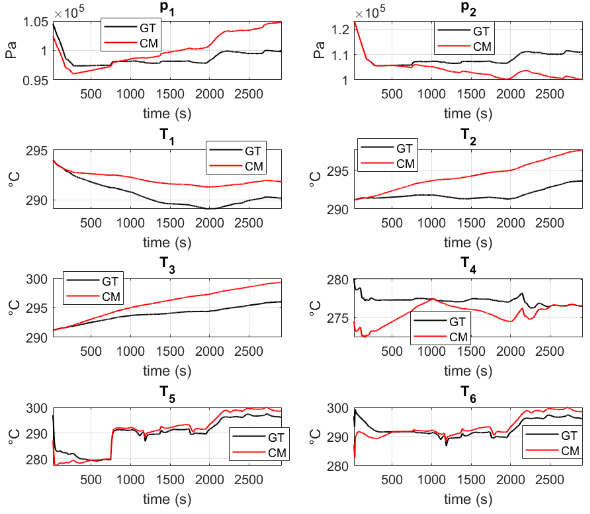}
\caption{ Comparison of the state evolution $\mathbf{Z}(t) = [ p_1,  p_2, T_1, T_2, T_3, T_4, T_5, T_6]$ for a given flight between the Coarse model (CM) and the Ground truth (GT)}
\label{fig:GTvsCM}
\end{figure}

\begin{figure}
\centering
\includegraphics[width=\linewidth]{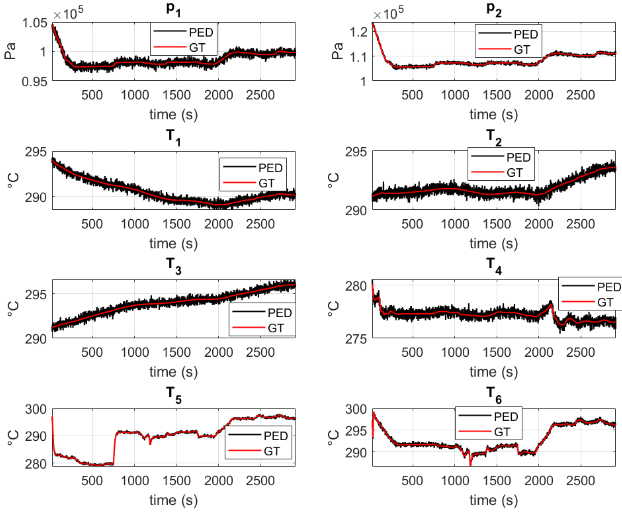}
\caption{ Comparison of the system evolution, $\mathbf{Z}(t) = [ p_1,  p_2, T_1, T_2, T_3, T_4, T_5, T_6]$, for a given flight between the Ground truth (GT) and the pseudo-experimental (noisy) data, PED}
\label{fig:GTvsPED}
\end{figure}

Additionally, in Figures \ref{fig:GTvsCM} and \ref{fig:GTvsPED}, a comparison is shown between the three {types of data (CM, GT and PED)} for a given flight simulation. 

At this point, three different approaches were tested:
\begin{itemize}
\item Extracting a model from scratch from GT data (Section \ref{Sec:CM}).
\item Obtaining a model from scratch using the noisy pseudo-experimental data (PED) (Section \ref{Sec:PSRDN}).
\item Extracting a correction {term to enrich  the CM}, thus constructing the Hybrid Twin (Section \ref{Sec:HT}).
\end{itemize}

Advantages and weaknesses of each approach will be discussed. The mathematical details {of the CM anf GT models} are omitted for confidentiality reasons. However, this is not important for presenting, discussing and employing the proposed methodology and, moreover, a successful outcome will be a sign that the proposed approach can address current industrial needs.

\subsection{Extracting a model from scratch using the GT data} \label{Sec:CM}

\subsubsection{Procedure and results} \label{Sec:CM_procedure}

It is interesting to analyse whether the proposed approach presented in Section \ref{Sec:Stable} is able to learn a model from scratch employing the GT data. Therefore, this section is focused on this goal. In next sections, we will analyse if it is able to obtain similar results when learning in the presence of noise (PED data) and finally, we will see the advantages of using the hybrid twin rational instead of the complete data-driven approach.

\begin{figure}
\centering
\includegraphics[width=0.9\linewidth]{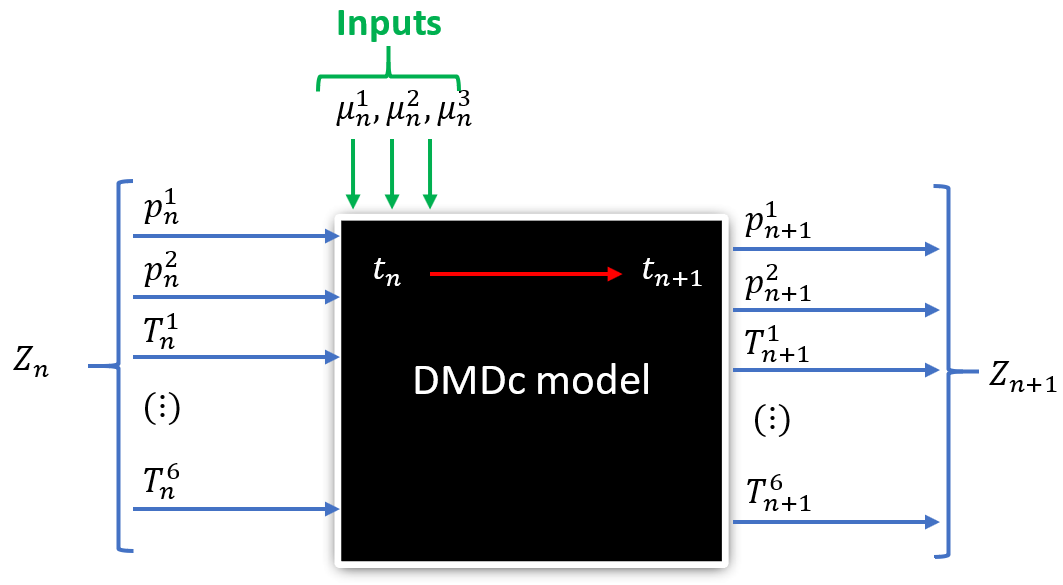}
\caption{ Diagram illustrating the inputs and the state vector of the proposed DMDc model to reproduce the pruned data of the system }
\label{BBox}
\end{figure}

We sketch the technique in the diagram of Figure \ref{BBox}. As it can be noticed, the system is characterized by eight variables defining the state of the system  (six temperatures $T^i_n$, $i=1, 2, \ldots, 6$, and two pressures $p^j_n$, $j=1,2$) and three control variables $\mu^k_n$, $k=1, 2, 3$ for each time instant $n$.

In the present case, we consider the simplest modeling approach in which the model consists in a simple linear application that maps the present state and control parameters $\mathbf Z_n$ and $\boldsymbol \mu_n$, respectively, onto the next system state $\mathbf Z_{n+1}$.

The available data consist in $\mathtt F=82$ flights, each one leading to eight time series $\mathbf Z_n^\mathtt f$, with $\mathtt f = 1, \dots, \mathtt F$, and $n=1, \dots , \mathtt n^\mathtt f$ (the number of collected data depends on the flight, these having different duration). It is important to note that, for any flight $t_{n+1}-t_n = \Delta t$, with constant $\Delta t$, for any component state and any flight.

We define the extended state as shown in Section \ref{Sec:Stable}:
\begin{equation}
\tilde{\mathbf Z}_n = \left ( 
\begin{array}{c}
\mathbf Z_n \\
\boldsymbol \mu_n
\end{array}
\right ).
\end{equation}

To learn the model, we select arbitrarily two flights from the $\mathtt F$ available, $\mathtt f= \mathtt r$ and $\mathtt f= \mathtt s$, and define the training matrices
\begin{equation}
{\tilde{\mathbf X}_0} = \left ( 
\begin{array}{cccccc}
\mathbf Z_0^\mathtt r & \cdots & \mathbf Z_{\mathtt n^\mathtt r}^\mathtt r & \mathbf Z_0^\mathtt s & \cdots & \mathbf Z_{\mathtt n^\mathtt s -1}^\mathtt s \\
\boldsymbol \mu_0^\mathtt r & \cdots & \boldsymbol \mu_{\mathtt n^\mathtt r}^\mathtt r & \boldsymbol \mu_0^\mathtt s & \cdots & \boldsymbol \mu_{\mathtt n^\mathtt s - 1}^\mathtt s
\end{array}
\right ),
\end{equation}
and 
\begin{equation}
{\mathbf{X}_1} = \left [
\begin{array}{cccccc}
\mathbf Z_1^\mathtt r & \cdots & \mathbf Z_{\mathtt n^\mathtt r}^\mathtt r & \mathbf Z_1^\mathtt s & \cdots & \mathbf Z_{\mathtt n^\mathtt s}^\mathtt s \\
\end{array}
\right ],
\end{equation}
that allows extracting the model by solving the problem indicated in Eq. \eqref{eq:ridge3} or its reduced counterpart \eqref{eq:ridgerom}.

Then, as soon as the model is extracted, we proceed to predict the state evolution for each one of the $\mathtt F$ flights, from their initial states, by simply writing at each time $t_n$, $n = 1, \dots , \mathtt n^\mathtt f$,
\begin{equation}
\tilde{\mathbf Z}_n^\mathtt f = \left ( 
\begin{array}{c}
\mathbf Z_n^\mathtt f \\
\boldsymbol \mu_n^\mathtt f
\end{array}
\right ),
\end{equation}
and applying the updating of Equations \eqref{eq:DS_wI} or \eqref{eq:ridgerom}. Training data are composed of two flights while the other eighty are used to test the performance of the present approach.

\begin{figure}
\centering
\includegraphics[width=\linewidth]{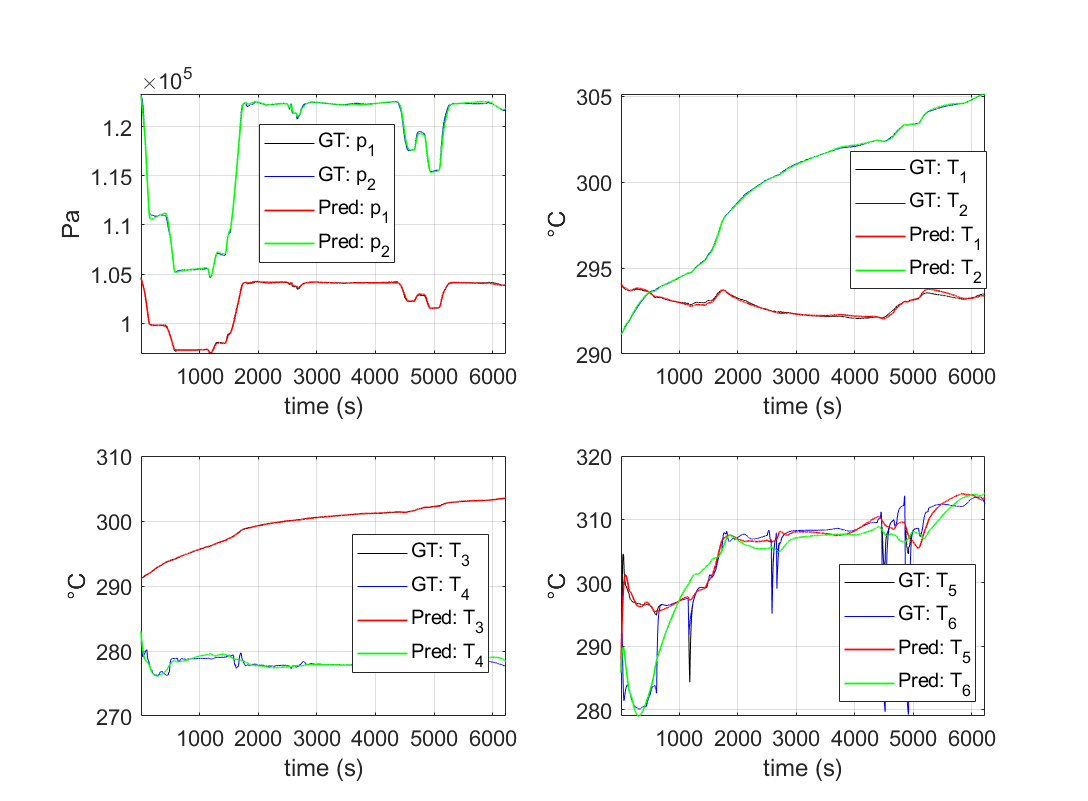}
\caption{ {Prediction of the GT data using the proposed technique for a flight which is not used in the training set. ``GT" refers to GT data series described in Section \ref{Sec:typedata} and ``Pred" refers to the stabilized DMDc model obtained with the proposed approach discussed in Section \ref{Sec:Stable}} }
\label{fig:CM_TSET}
\end{figure}

Figure \ref{fig:CM_TSET} compares the predicted states at each time instant by integrating the just unveiled model from the initial condition. It employs a GT data series corresponding to one particular, previously unseen flight. It can be observed that the proposed approach achieves an excellent agreement for variables $p_1$, $p_2$, $T_1$, $T_2$, $T_3$ and $T_4$.

\begin{figure}
\centering
\includegraphics[width=0.9\linewidth]{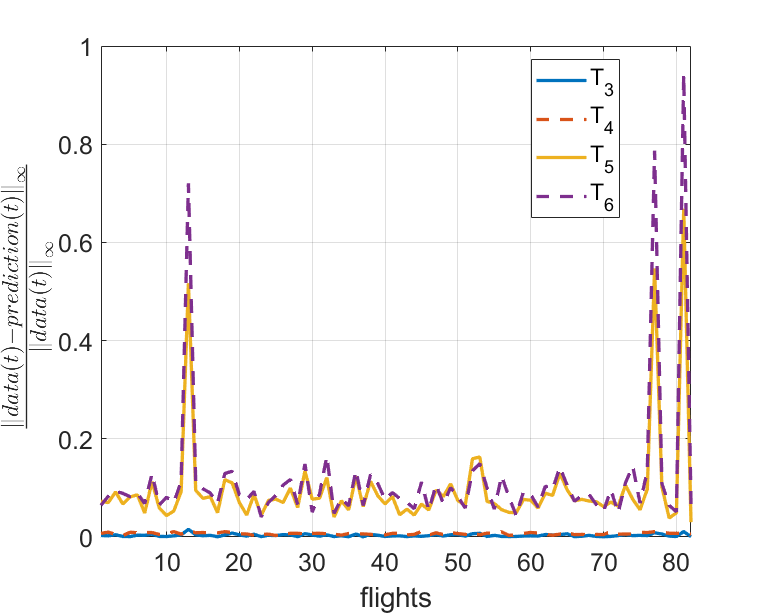}
\caption{ {Error in the prediction of $T_3$, $T_4$, $T_5$ and $T_6$ for different flights which are not  in the training set. The prediction error in variables $T_5$ and $T_6$ is higher than the other ones due to their fast time evolution}}
\label{fig:compO5toO8}
\end{figure}

On the other hand, although the error in variables $T_5$ and $T_6$ is larger it achieves to follow the general trend, despite the fast time evolutions that these variables exhibit. The same tendency is observed in all the flights as  in Fig. \ref{fig:compO5toO8}) proves.
To better capture the fast evolutions of these two variables, the procedure described in Section \ref{M1} for addressing nonlinear behaviors \citep{sancarlos2020rom} could be employed. 
However, in this work it is preferred to improve the accuracy in the prediction of these special variables by employing a Hybrid approach, for the reasons exposed in Section \ref{Sec:HT}.

\begin{figure}
\centering
\includegraphics[width=0.9\linewidth]{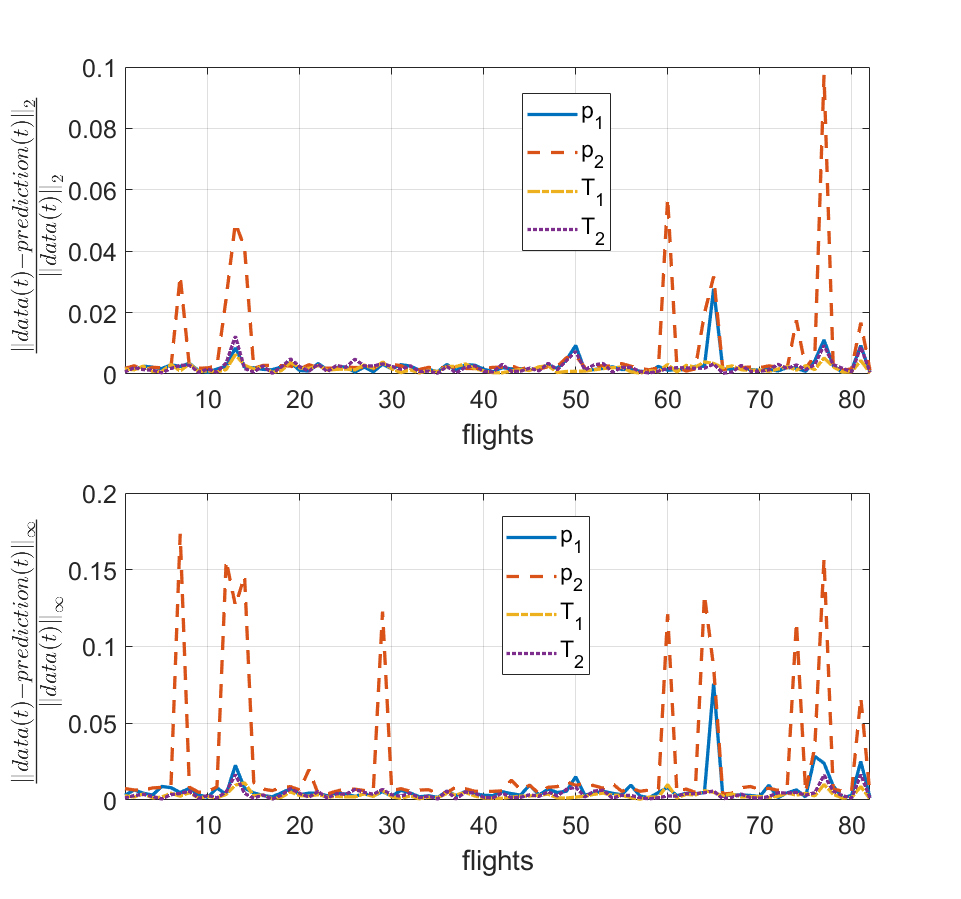}
\caption{ {Error in the prediction of $p_1$, $p_2$, $T_1$ and $T_2$ of the proposed technique for lights which are not in the training set} }
\label{fig:GcompO1to4}
\end{figure}

\begin{figure}
\centering
\includegraphics[width=0.9\linewidth]{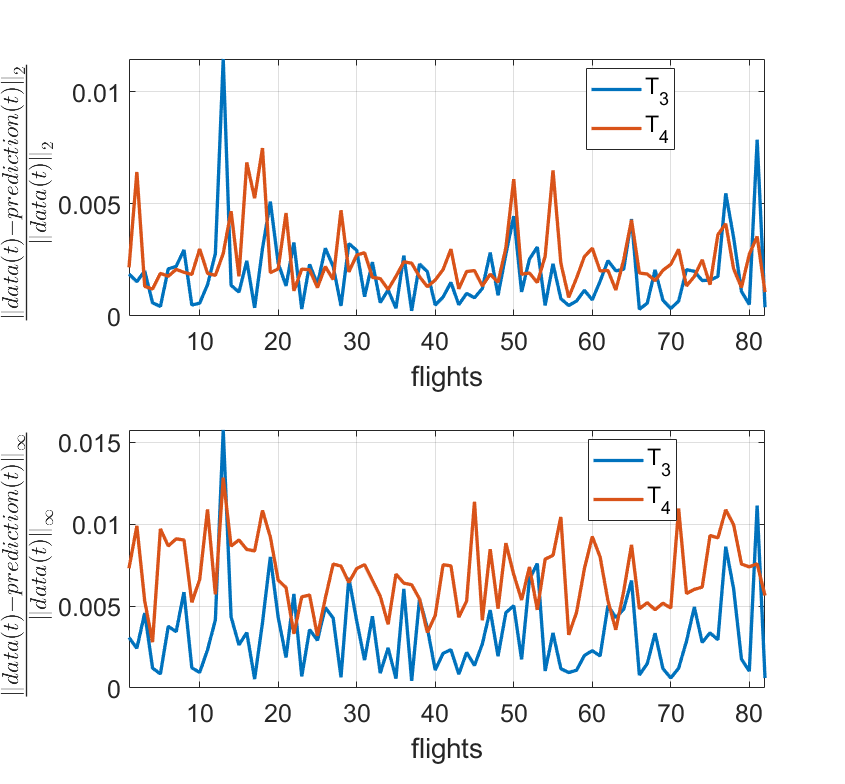}
\caption{ {Error in the prediction of $T_3$ and $T_4$ of the proposed technique for flights which are not  in the training set} }
\label{fig:GcompO5to6}
\end{figure}

In Figure \ref{fig:GcompO1to4}, it is observed that a similar accuracy is obtained  for more than 85 \% of the flights in the testing set (concerning variables $p_1$, $p_2$, $T_1$ and $T_2$). A similar conclusion follows from Figure \ref{fig:GcompO5to6} for variables $T_3$ and $T_4$. Therefore, it is concluded that the model has a good ability for generalization taking into consideration that just two flights are considered in the training. 

However, the error in variables $T_5$ and $T_6$ can reach high values in a considerable part of the testing set. To adress that, the HT rationale will be proposed and discussed.

Before that, we will see what happens if stability is not enforced when extracting the GT model from scratch in the following subsection \ref{Sec:improvements} as well as the improvements of the proposed procedure.


\subsubsection{Checking explicitly the improvements of the proposed approach to learn stable systems. Example when proceeding from the GT data.} \label{Sec:improvements}

In the above Section, the results when modelling the GT data were shown when using the stabilization procedure for the DMDc proposed in this work. However, it is interesting to analyze whats happens when using other algorithms or by simply using standard procedures to observe the benefits of the proposed stabilization.

\begin{figure}
\centering
\includegraphics[width=0.9\linewidth]{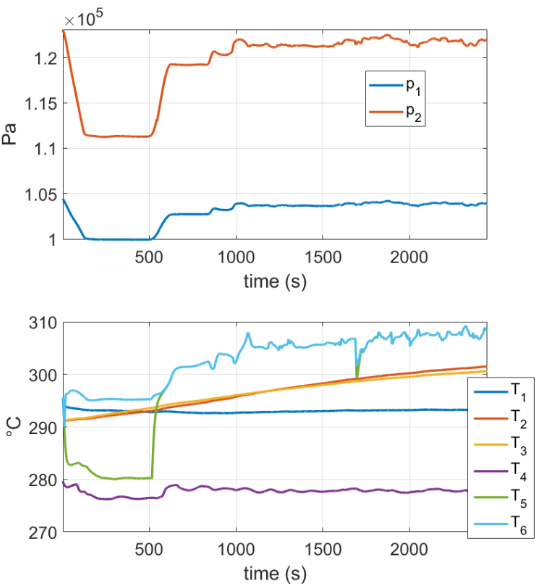}
\caption{ GT data for the example in Section \ref{Sec:improvements}}
\label{fig:CMflight}
\end{figure}

To this end, the GT data for a particular flight is considered. The time evolution of the system can be observed in Figure \ref{fig:CMflight}. 

The first objective is to see if we can approximate the dynamics of Figure \ref{fig:CMflight} by constructing a standard DMDc model without stabilizing it. Moreover,  we are interested in observing if stability issues arise.

\color{black}
Therefore,  a model is constructed following the DMDc procedure with the state vector and inputs exposed in Section \ref{Sec:CM_procedure}. 

As shown in Figure \ref{fig:DMDcflight} below, the {DMDc model} reports stability issues {giving useless predictions}. 

\begin{figure}
\centering
\includegraphics[width=0.9\linewidth]{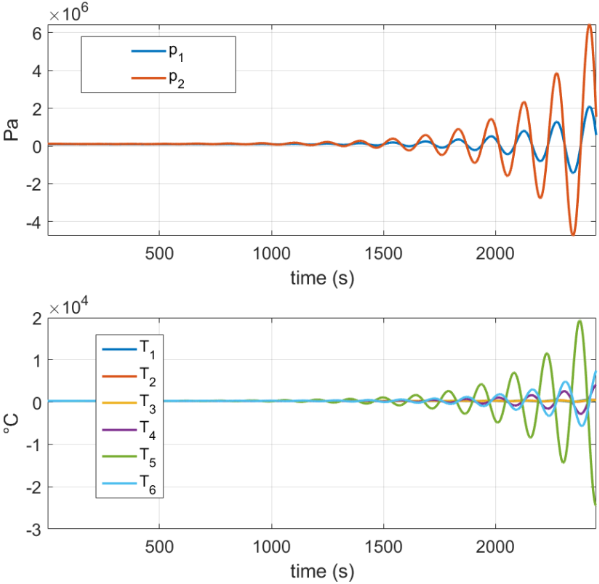}
\caption{ Prediction $\mathbf{Z}(t) = [ p_1,  p_2, T_1, T_2, T_3, T_4, T_5, T_6]$ of the GT obtained through DMDc. This prediction tries to reproduce the flight of Figure \ref{fig:CMflight} but fails to provide with stable results}
\label{fig:DMDcflight}
\end{figure}

\begin{figure}
\centering
\includegraphics[width=\linewidth]{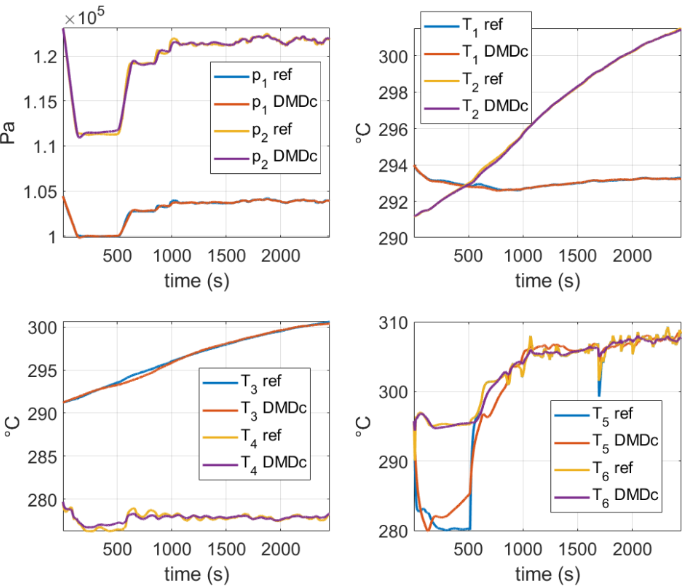}
\caption{ { Comparison between the reference dynamics of Figure \ref{fig:CMflight} and the prediction of the modified, stable DMDc model. Huge improvements are observed when comparing with Figure \ref{fig:DMDcflight}. The state vector of the system is  $\mathbf{Z}(t) = [ p_1,  p_2, T_1, T_2, T_3, T_4, T_5, T_6]$} }
\label{fig:ridgeDMDcflight}
\end{figure}

Nevertheless, by applying the proposed stabilization approach to the DMDc algorithm {(as discussed in Section \ref{Sec:Stable})}, a stable model is easily obtained quick and fast. The flight predicted by the proposed approach is shown in Figure \ref{fig:ridgeDMDcflight}. It is worth noting the great improvement observed by comparing { the standard non-stabilized DMDc model (Figure \ref{fig:DMDcflight}) with the stabilized one (Figure \ref{fig:ridgeDMDcflight}) able to capture complex dynamics while completely overcoming the stability issues.} 

Now, in the next section, we will examine the possibility of unveiling an accurate model from noisy data. In this way, two important aspects will be analysed: the interest of employing a Hybrid Twin approach and the ability of the proposed technique to filter noise.

\subsection{Extracting a model from scratch using noisy data (PED).} \label{Sec:PSRDN}

In this section, we attempt to unveil a model from scratch using noisy data. This will allow us to study the robustness of the approach in the filtering process.

After applying the technique in different flights and studying the reconstruction error by considering different extended states $\tilde{\mathbf Z}$, the proposed model is composed of:
\begin{equation}
\tilde{\mathbf Z}_n = \left ( 
\begin{array}{c}
\mathbf Z_n \\
\boldsymbol \mu_n \\
\boldsymbol \mu_{n-1} \\
\boldsymbol \omega_{n} \\
\mathbf W_n
\end{array}
\right ),
\end{equation}
where:
\begin{equation*}
\boldsymbol \omega_{n} = \sum_{i=0}^n \mu_{i} (t_{i+1}-t_i),
\end{equation*}
and 
\begin{equation*}
\mathbf W_{n} = \sum_{i=0}^n \boldsymbol \omega_{i} (t_{i+1}-t_i).
\end{equation*}

The same methodology of the previous Section is applied, by using the new extended states $\tilde{\mathbf Z}$. This way, we can address the more complex behavior of the noisy data. In this case, nine flights of different duration are used for the training set.

\begin{figure}
\centering
\includegraphics[width=\linewidth]{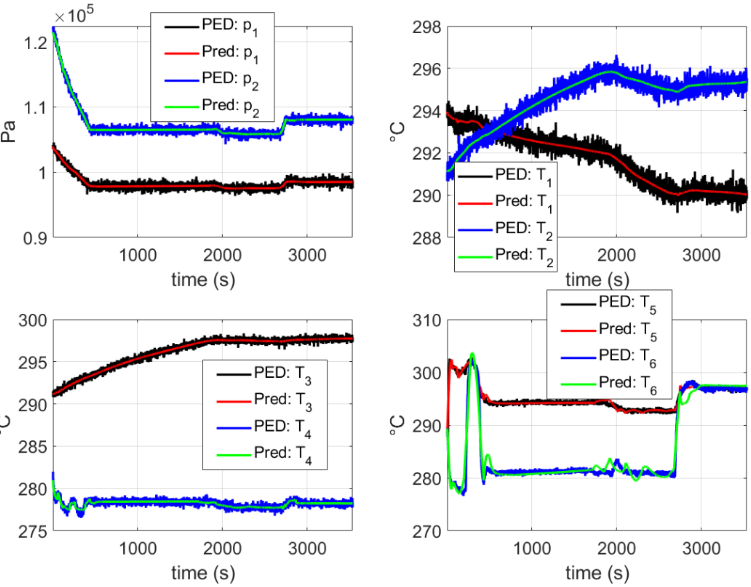}
\caption{ {Comparison between the model obtained from scratch using PED data and the PED data itself. In this figure, the reconstruction of a flight contained in the training set is shown. ``PED" refers to the pseudo-experimental data with noise described in Section \ref{Sec:typedata} and ``Pred" refers to the stabilized DMDc model obtained with the proposed approach discussed in Section \ref{Sec:Stable}. It can be observed that an excellent agreement is obtained for every variable while filtering the noise} }
\label{fig:PSRDN_TS}
\end{figure}

In Fig. \ref{fig:PSRDN_TS}, a comparison is shown between the dynamics predicted by the DMDc model obtained from scratch using the PED and the PED itself (see Section \ref{Sec:typedata}), for a flight contained in the training set. Here, it is observed that the model can capture the dynamics of the system with an excellent accuracy, by just employing the initial state of the system and the corresponding control inputs while filtering the noise contained in the training data. 

\begin{figure}
\centering
\includegraphics[width=\linewidth]{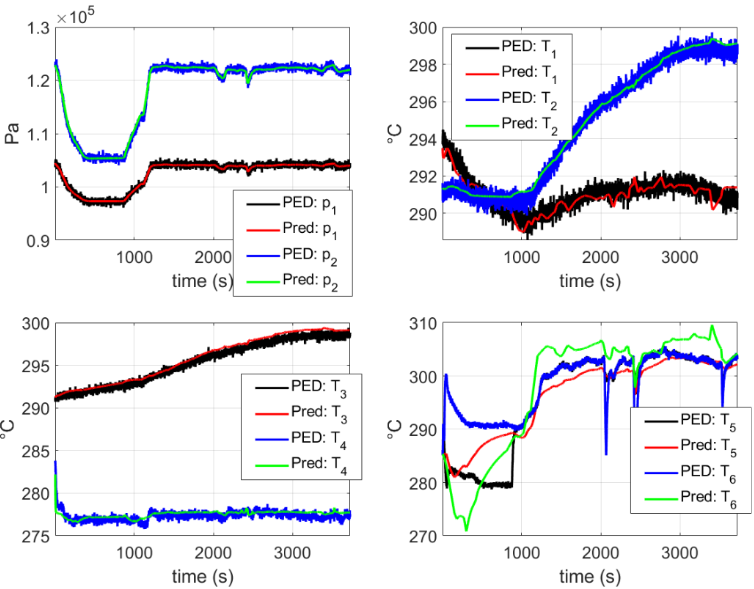}
\caption{ {Comparison between the model obtained from scratch using PED data and the PED data itself. In this figure, the reconstruction of a flight which is not contained in the training set is shown. ``PED" refers to the pseudo-experimental data with noise described in Section \ref{Sec:typedata} and ``Pred" refers to the stabilized DMDc model obtained with the proposed approach discussed in Section \ref{Sec:Stable}. It can be observed that a good agreement is obtained for all the variables with the exception of $T_5$ and $T_6$ 
} }
\label{fig:PSRDN_US}
\end{figure}

In Fig. \ref{fig:PSRDN_US} a comparison is shown {between the dynamics predicted by the DMDc model obtained from scratch using the PED and the PED itself (see Section \ref{Sec:typedata})}, for a flight contained in the testing set (never considered in the model construction). In these plots, it is shown that a good agreement is obtained for all the variables with the exception of $T_5$ and $T_6$. Similar results are reported in the other flights of the testing set.

Variables $T_5$ and $T_6$ are more challenging to predict because of their fast time evolution. Nevertheless, we are going to deal with them in the next section. Therefore, the Hybrid Twin concept, which follows, is expected leading to more accurate results.

\subsection{Extracting the correction model. Hybrid Twin paradigm.}
\label{Sec:HT}

As discussed in Section \ref{M2}, constructing a model of the real system from scratch is not the most valuable route when addressing complex systems. For the system analyzed in the present work, outputs $p_1$, $p_2$, $T_1$, $T_2$, $T_3$ and $T_4$ are predicted to a great accuracy for the approaches shown in Sections \ref{Sec:CM} and \ref{Sec:PSRDN}. On the other hand, outputs $T_5$ and $T_6$ present difficulties. 
In these cases, an interesting option consists in expressing the state of the system from an additive correction of the coarse model. Therefore, in this case, the proposed model is going to capture just the ignorance that the coarse model contains. 

One of the advantages of this concept is that the main response is provided by the physics-based model, thus guaranteeing that the model is going to exhibit a behavior coherent with the physical phenomenon under scrutiny as well as being explained by practitioners. In addition, the part of the response which has difficulties in being modeled---for instance, the appearance of degradation of the system---can be approximated by the data-driven model.

The Hybrid Twin (HT) concept is illustrated in Figure \ref{HTconcept}. Note that only the first measurement is mandatory to run what we coined as the $\Delta \mathbb M$. Therefore, knowing the initial state of the system, the real response can be reproduced adding the correction model to the CM without further measurements. 


Again, the extended state $\Delta \tilde{ \mathbf Z}$ for the discrepancy model is:
\begin{equation}
\Delta  \tilde{\mathbf Z}_n = \left ( 
\begin{array}{c}
\Delta \mathbf Z_n \\
\boldsymbol \mu_n \\
\boldsymbol \mu_{n-1} \\
\boldsymbol \omega_{n} \\
\mathbf W_n
\end{array}
\right ),.
\end{equation}

Nine flights are considered in the training set. As expected for a real-life application, the measured data (that is, the PED) is employed to obtain the discrepancy to be modeled within the HT concept.

\begin{figure}
\centering
\includegraphics[width=\linewidth]{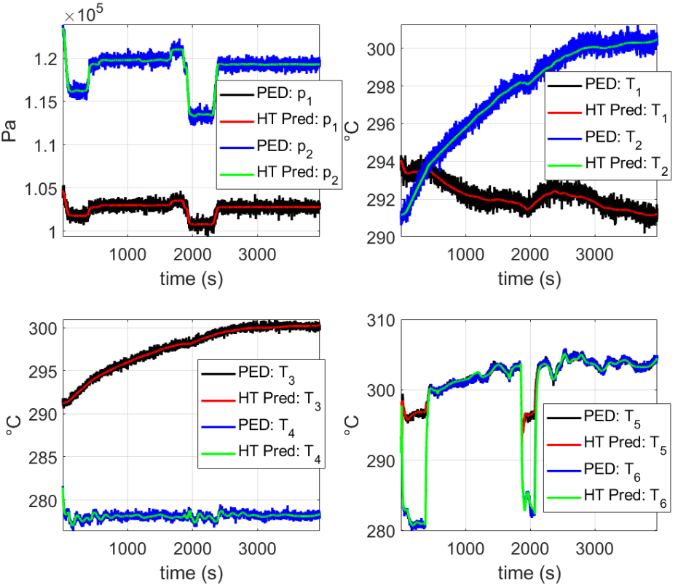}
\caption{ {Prediction of the HT approach considering a flight in the testing set. ``PED" refers to the pseudo-experimental data with noise described in Section \ref{Sec:typedata} and ``HT Pred" refers to the HT approach whose correction term corresponds to a stabilized DMDc model obtained with the methodology discussed in Section \ref{Sec:Stable}. The correction term was constructed using the PED. It can be observed that an excellent agreement is obtained for all the variables}\label{fig:HT_pred} }

\end{figure}

\begin{figure}
\centering
\includegraphics[width=\linewidth]{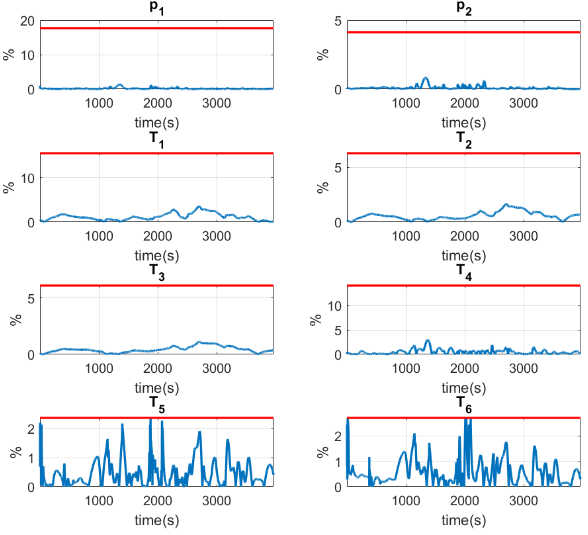}
\caption{ {Error of the HT approach (blue line) considering a flight which is not used for the training. The red line refers to the maximum error in the pseudo measurements (PED).  The error criterion is defined in Eqs (\ref{err1}) (blue line) and (\ref{err2}) (red line) }}
\label{fig:HT_error}
\end{figure}

\begin{figure}
\centering
\includegraphics[width=\linewidth]{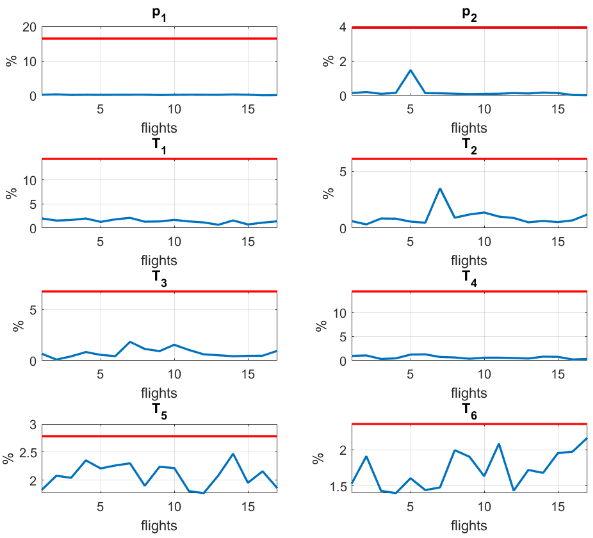}
\caption{ {Error of the HT approach (blue line) considering different flights which are not used for the training. The red line refers to the maximum error  in the pseudo measurements. The error assigned to a flight is the mean value of the error defined in Eq. (\ref{err1})}  }
\label{fig:HT_errTESTSET}
\end{figure}

Figures \ref{fig:HT_pred}, \ref{fig:HT_error} and \ref{fig:HT_errTESTSET} are obtained by just integrating from the initial state of the system and by employing {and enriching} the CM prediction without any further measurement. This proves the excellent agreement that can be achieved within the HT rationale.

In Fig. \ref{fig:HT_pred}, a comparison is shown between the HT prediction and the PED for a previously unseen flight. In these plots, an excellent agreement is noticed for all the variables. Moreover, predictions filter the noisy measurements.

An error criterion is defined to compare the prediction of the HT approach with the accuracy of the measuring instruments:
\begin{equation} \label{err1}
\text{err}_i(t) = \frac{z_i^{\text{GT}}(t)  - z_i^{{HT}}(t)}{ \Delta z_i^{\text{max}, \text{GT} }} ,
\end{equation}
\begin{equation} \label{err2}
\text{err}^{\text{max}}_{\text{meas}} = \frac{ \max(z_i^{\text{GT}}  - z_i^{m} )} { \Delta z_i^{\text{max}, \text{GT} }},
\end{equation}
where:
\begin{itemize}
\item $i$ refers to the $i$-th variable $z_i$ of the state vector.
\item $z_i^{{HT}}$ is the predicted value of the HT for $z_i$. 
\item $z_i^{m}$ is the measured value of $z_i$. These data include the corresponding noise. In other words, these data are the PED.
\item $z_i^{\text{GT}}$ is the true value of $z_i$ which is theoretically unknown and cannot be accessed by an observer in a real application. We use this value for evaluation purposes.
\item $\Delta z_i^{\text{max}, \text{GT}}$ is the difference between the maximum and the minimum value of $z_i^{\text{GT}}$ considering all the flights.
\end{itemize}

\begin{figure}
\centering
\includegraphics[width=\linewidth]{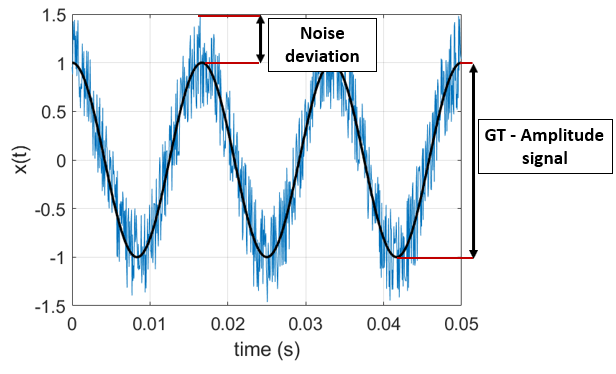}
\caption{ {Sine wave with noise. In the plot, the maximum variation of the signal is indicated as well as the deviation caused by the noise to illustrate the concept used to define the error criterion} }
\label{fig:signoise}
\end{figure}

Using Eqs. \eqref{err1} and \eqref{err2}, it is possible to compare the accuracy of the HT with the one obtained by measuring the data (that is, the PED). Moreover, the relative errors are computed taking as a base the maximum variation of each variable when regarding all the available flights. Fig. \ref{fig:signoise} shows the maximum variation of a signal as well as the deviation caused by the noise to illustrate the concept.

Observing Figs. \ref{fig:HT_error} and \ref{fig:HT_errTESTSET}, we can confirm that the HT concept allow us to improve, not only the accuracy of the CM but the accuracy of the measuring instruments regardless of whether or not flights come from training by filtering the noise. In this way, two goals are achieved at the same time: enriching the CM by learning the difference with the measured data while filtering the noise.

Therefore, the HT concept is a valuable route to enrich  physics-based models with data-driven corrections. It is important to note that for HT to be applied, the CM must not be extremely bad, since in this case a direct data-driven or reduced-order modeling approach would be preferred (because there is no point in correcting such a model).


%

\section{Conclusions} \label{Sec:Conclusions}

This work presents a fast and efficient methodology, covering several variants to learn dynamical models while guaranteeing a low computational cost as well as the achievement of stable dynamical time integrations. This technique was used with success to predict a practical scenario under the HT rationale, being able to impose stability in the correction term.
In addition, the proposed technique filters noise improving the knowledge of the system state.

We also compared the proposed technique in two scenarios: \textit{when it is employed to obtain models from scratch} versus \textit{when it is employed for an enrichment in the HT rationale}. 

We concluded that for more complex systems the HT paradigm seems advantageous for two reasons. The first one is that more complex behaviors can be captured (as variables $T_5$ and $T_6$). The second one is that, in the HT, the main response is relied on the physics-based model thus guaranteeing that the model is going to exhibit a behavior coherent with the physical phenomenon at hand. Consequently, just the part of the response which has difficulties in being modeled is carried out by the data-driven model, for instance, degradation or aging.
\color{black}

\end{document}